\documentclass[aps,pra,reprint,longbibliography]{revtex4-1}

\usepackage{graphicx}
\usepackage{amsmath}
\usepackage{amssymb}
\usepackage{textcomp}

\begin{document}
	
\title[]{Superconducting Optoelectronic Neurons \uppercase\expandafter{\romannumeral 3 \relax}: Synaptic Plasticity}

\author{Jeffrey M. Shainline, Adam N. McCaughan, Sonia M. Buckley, Christine A. Donnelly, Manuel Castellanos-Beltran, Michael L. Schneider, Richard P. Mirin, and Sae Woo Nam}
\affiliation{National Institute of Standards and Technology, 325 Broadway, Boulder, CO, 80305}
			
\date{\today}
	
\begin{abstract}
As a means of dynamically reconfiguring the synaptic weight of a superconducting optoelectronic loop neuron, a superconducting flux storage loop is inductively coupled to the synaptic current bias of the neuron. A standard flux memory cell is used to achieve a binary synapse, and loops capable of storing many flux quanta are used to enact multi-stable synapses. Circuits are designed to implement supervised learning wherein current pulses add or remove flux from the loop to strengthen or weaken the synaptic weight. Designs are presented for circuits with hundreds of intermediate synaptic weights between minimum and maximum strengths. Circuits for implementing unsupervised learning are modeled using two photons to strengthen and two photons to weaken the synaptic weight via Hebbian and anti-Hebbian learning rules, and techniques are proposed to control the learning rate. Implementation of short-term plasticity, homeostatic plasticity, and metaplasticity in loop neurons is discussed.  
\end{abstract}
	
	
\maketitle
	
\section{\label{sec:introduction}Introduction}
Information processing systems with differentiated processing, information integration, and distributed memory modeled after biological neural systems are appealing as tools for understanding neural and nonlinear dynamical systems as well as for computation in contexts requiring complex contextualization and dynamic learning. In such neural systems \cite{daab2001,geki2002}, the synaptic weights between nodes in the network are crucial memory elements that affect dynamics and computation \cite{abre2004,bu2006,siqu2007,haah2015}. For some applications, it is important to have a means by which a user can interface with the system to externally control the synaptic weights to implement learning algorithms \cite{ni2015}. In other applications, it is desirable for the synaptic weights to dynamically update based on network activity in an unsupervised manner. It is beneficial for a hardware platform to be capable of both. 

It has been proposed \cite{shbu2017,sh2018a} that combining the strengths of light for communication and superconducting electronics for efficient computation offers a route to large-scale neural systems. A circuit that transduces single-photon communication signals to integrated supercurrent has been described in Ref. \onlinecite{sh2018b}. In that reference, a means to modify the synaptic weight via a bias current ($I_{\mathrm{sy}}$) was identified. The present work explores circuits that dynamically control $I_{\mathrm{sy}}$. 

The circuits implemented to control $I_{\mathrm{sy}}$ should meet several criteria: 1) Transition between the minimum and maximum values of $I_{\mathrm{sy}}$ should be possible with a specified number of increments to control the learning rate; 2) The circuit should not be able to set $I_{\mathrm{sy}}$ outside of this range so that simple update rules or training algorithms do not result in excessively large synaptic weights; 3) It should be possible to cycle the value of $I_{\mathrm{sy}}$ from minimum to maximum and back repeatedly without degradation; 4) In addition to a means by which the synaptic weights can be incremented by an external supervisor, there should be a means by which correlated photon signals from the two neurons associated with a synapse can strengthen or weaken the synaptic weight depending on the relative arrival times of the signals from the two neurons; 5) Within this unsupervised mode of operation, synaptic update events should be induced by single-photon signals to fully exploit the energy efficiency of the superconducting optoelectronic hardware; 6) The transition probability between synaptic states should also be dynamically adjustable based on photonic signals to achieve metaplastic behavior. This paper explores circuit designs satisfying all these criteria.

\begin{figure} 
	\centerline{\includegraphics[width=8.6cm]{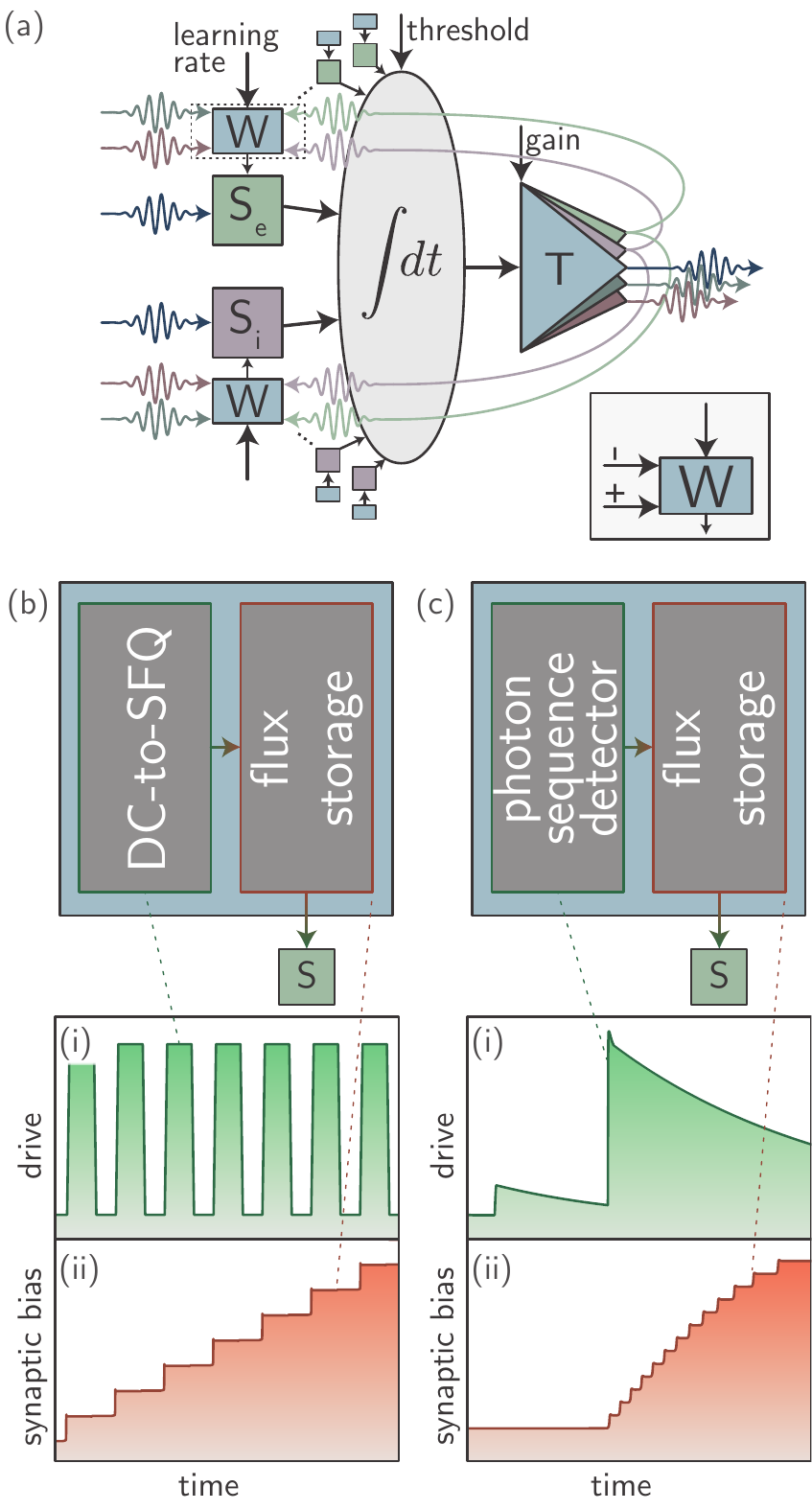}}
	\caption{\label{fig:synapticPlasticity_schematic}(a) Schematic of the neuron showing excitatory ($\mathsf{S_e}$) and inhibitory synapses ($\mathsf{S_i}$) connected to an integration loop with a variable threshold. The wavy, colored arrows are photons, and the straight, black arrows are electrical signals. See Ref.\,\onlinecite{sh2018b} for details. The dashed rectangle labels the synaptic update box ($\mathsf{W}$), which is the focus of this work. The number of fluxons is determined by the synaptic bias current, which is controlled by $\mathsf{W}$. The photons from the left and right accomplish spike-timing-dependent plasticity. The inset at the lower right depicts the electrical signals used for supervised learning. (b) Synaptic update in supervised mode. Square pulses add or remove fluxons from a loop, which strengthens or weakens the synaptic weight. (c) Synaptic update in unsupervised mode. Photons from pre-synaptic and post-synaptic neurons are used to change the flux in the loop.}
\end{figure}

A schematic of the neuron under consideration is shown in Fig.\,\ref{fig:synapticPlasticity_schematic}(a). Operation is as follows. Photons from afferent neurons are received by single-photon detectors (SPDs) \cite{gook2001,nata2012,liyo2013,mave2013,veko2015} at a neuron's synapses. Using Josephson junctions (JJs) \cite{ti1996,vatu1998,ka1999}, these detection events are converted into an integrated supercurrent that is stored in a superconducting loop. The amount of current added to the integration loop during a synaptic photon detection event is determined by the synaptic weight. The synaptic weight is dynamically adjusted by another circuit combining SPDs and JJs. When the integrated current from all the synapses of a given neuron reaches a threshold, an amplification cascade begins, resulting in the production of light from a waveguide-integrated LED. The photons thus produced fan out through a network of passive dielectric waveguides and arrive at the synaptic terminals of other neurons where the process repeats.

The dashed box in Fig.\,\ref{fig:synapticPlasticity_schematic}(a) encloses the synaptic weight control circuits that are the focus of this work. These signals add or remove flux from a storage loop, which is inductively coupled to the current bias line, $I_{\mathrm{sy}}$. This loop is referred to as the synaptic storage (SS) loop (Fig.\,\ref{fig:synapticPlasticity_binaryCircuit}), and the flux stored in this loop functions as the memory for the synapse.

The circuits described in this work modify $I_{\mathrm{sy}}$ in either a supervised manner using JJs  or unsupervised manner using SPDs in conjunction with JJs. Qualitative explanation of the memory update process in shown in Fig.\,\ref{fig:synapticPlasticity_schematic}(b) for supervised learning and in Fig.\,\ref{fig:synapticPlasticity_schematic}(c) for unsupervised learning. For the simplest binary synapse, a flux-quantum memory cell can be used to switch between the strong and weak synaptic states in 50\,ps. This binary design can be extended to a multi-stable synapse that can modify the synaptic weight between the fully potentiated and fully depressed states with hundreds of stable intermediate levels, and implementations with more or less resolution are straightforward to achieve. For unsupervised learning, we consider a circuit that can implement a Hebbian learning rule that potentiates a synaptic connection using one photon from the pre-synaptic neuron and one photon from the post-synaptic neuron. We generalize this circuit to implement full spike-timing-dependent plasticity (STDP) wherein a synaptic weight can be either potentiated or depressed based on Hebbian and anti-Hebbian timing correlations. This STDP circuit uses single-photon signals at four ports. Implementations of short-term plasticity, homeostatic plasticity, and metaplasticity are discussed in the Appendices. Combining these synapse designs, it is possible to realize neurons with a distribution of synapses that update at different rates as well as ensembles of neurons wherein different neurons store information about different stimuli learned at different times, thus achieving a network with rapid adaptability and long memory retention times necessary for cognition, as discussed in Ref.\,\onlinecite{sh2018a}.   
	
\section{\label{sec:conceptualOverview}Conceptual overview}
The field of neural computing \cite{scpo2017,lide2015} is broad and deep. A rich body of work exists wherein neural concepts are implemented in software using conventional Boolean hardware \cite{ho1982,achi1985,bi1994,hisa2006}. Such technologies are usually referred to as neural networks. We make the distinction that neural computing utilizes hardware with neural behavior present in the physics of the devices that implement the required computations rather than implementing the neural computational functions with algorithms in software. Software neural nets and neural hardware are both useful and both have promise to affect the advanced computing landscape in coming years.

We further delineate two main modes of operation of neural computers. In one mode, controlled inputs are presented to the system, and the system provides an output. The output from the system is compared to a desired output, and an error is calculated based on a cost function. This error is then used to update the configuration of the system, often through backpropagation \cite{ni2015}. We refer to this mode of operation as ``supervised learning''. Most technologies commonly referred to as machine learning or deep learning operate in this mode. The objective of supervised learning is often to train the hardware to perform a specific task \cite{sihu2016}. 

For larger neural systems performing general cognitive functions, it is advantageous to operate in an unsupervised manner. In supervised learning, there is an external means by which properties affecting network operation can be adjusted (such as by explicitly changing synaptic weights or neuron thresholds). In unsupervised learning, no such external control is available. Unsupervised learning is scalable in that the user is not required to calculate or adjust the network parameters, so systems with many more degrees of freedom can be realized. Yet unsupervised learning requires that internal activity of the network be capable of adjusting the degrees of freedom to form a useful representation of the information it is expected to process. Unsupervised learning usually occurs within spiking dynamical systems. In these dynamical systems, modification of the synaptic weights changes the structure of the network and therefore adapts the dynamical state space \cite{spto2000,budr2004} based on external stimulus and internal network activity.

In the present work, we are interested in both supervised and unsupervised modes of operation, and we focus on the means by which synaptic weights can be modified either externally or internally to enable training and learning. In unsupervised learning, we are interested in systems that will interact continuously with their environment, be capable of immediately assimilating new information, and also capable of remembering events as long as possible. Such competing memory demands are sometimes referred to as the adaptability-precision trade-off \cite{khso2017}, and the best-performing synapses in this regard are complex \cite{fudr2005} and may have many stable levels \cite{fuab2007}. In human subjects, memories have been observed to fade with a power law temporal dependence \cite{wieb1991,wieb1997}. It is difficult to do better than power law forgetting with plastic synapses that continually adapt \cite{fudr2005}, and simple synapses lose their memory trace most quickly \cite{fuab2007}. In the present work, we show synapses with a number of stable states ranging from two to hundreds. These synapses have dynamically variable memory update rates, making the synapses suitable for power law memory retention.

The neurons under consideration have been described in Ref.\,\onlinecite{sh2018b}, and they are referred to as loop neurons. They will be employed in the context of superconducting optoelectronic networks described in Refs.\,\onlinecite{shbu2017} and \onlinecite{sh2018a}. In such networks, light is used to communicate signals between neurons, and when a single photon is received at a synapse, the signal is converted to a number of flux quanta \cite{ti1996,vatu1998,ka1999}. We refer to this as a synaptic firing event. In such a neuron, the role of the synaptic weight is to change how many flux quanta are generated during a synaptic firing event. This determines how much current is added to the neuron's current integration loop \cite{sh2018b}, and therefore how close the neuron is to reaching threshold. When one or more synaptic firing events add sufficient current to the neuron's integration loop to reach threshold, the neuron is induced to produce a pulse of light, which is distributed to that neuron's downstream connections. Threshold detection and pulse production are treated in Ref.\,\onlinecite{sh2018d}. 

The objective of this paper is to describe the means by which the synaptic weight can be modified to enable dynamically reconfigurable synapses. The photon-to-fluxon transduction that occurs during a synaptic firing event is implemented with an SPD in parallel with a JJ, as described in Ref.\,\onlinecite{sh2018b}. To change the number of fluxons generated during the synaptic firing event, one can simply change the current bias across this JJ, referred to as the synaptic firing junction. The circuits presented here are designed to dynamically modify the current bias to the synaptic firing junction, $I_{\mathrm{sy}}$ (see Fig.\,2 of Ref.\,\onlinecite{sh2018b}). We refer to the circuits that modify $I_{\mathrm{sy}}$ as the synaptic update circuits. In general, there will be a chosen weakest synaptic strength and strongest synaptic strength at each synapse, and in general the weakest synaptic strength will be achieved with $I_{\mathrm{sy}}^{\mathrm{min}} > 0$. Thus, it is the goal of the synaptic update circuit to vary $I_{\mathrm{sy}}$ over some range $I_{\mathrm{sy}}^{\mathrm{min}}\le I_{\mathrm{sy}} \le I_{\mathrm{sy}}^{\mathrm{max}}$. In certain contexts it is sufficient for $I_{\mathrm{sy}}$ to only be able to take two values  \cite{lide2015}, while in other learning environments it may be advantageous to be able to achieve many values of $I_{\mathrm{sy}}$ between $I_{\mathrm{sy}}^{\mathrm{min}}$ and $I_{\mathrm{sy}}^{\mathrm{max}}$. Discussion of synapses for learning in various contexts is presented in Sec.\,\ref{sec:discussion}.

One means of modifying the current bias to the synaptic firing junction is depicted in Fig.\,\ref{fig:synapticPlasticity_binaryCircuit}. For systems with many neurons each with many synapses, we would like to use a single current source to establish the baseline synaptic bias to all synapses ($I_1$ in Fig.\,\ref{fig:synapticPlasticity_binaryCircuit}), keeping in mind that we may need the baseline synaptic bias to be different for different synapses. This can be achieved by using a single current bias, $I_1$, and using mutual inductors to couple this current to each synapse. The synaptic firing circuit is thus biased by a superconducting loop, referred to as the synaptic bias (SB) loop, and the objective of the synaptic update circuit is to change the current in the SB loop, also through mutual inductors. The circuits presented throughout this work achieve the various synaptic states by changing the amount of flux trapped in another superconducting loop, referred to as the synaptic storage (SS) loop. This basic concept is shown in Fig.\,\ref{fig:synapticPlasticity_binaryCircuit}, where the SB loop is coupled to both the main bias, $I_1$, and the dynamic synaptic bias based on the flux trapped in the SS loop. All circuits presented in the remainder of this work provide a means to adjust the flux stored in the SS loop.

To implement supervised learning, we would like to control the flux stored in the SS loop using simple control signals, which we take to be square current pulses. In Sec.\,\ref{sec:supervised} we show that such current pulses can be used to modify the flux in the SS loop and therefore $I_{\mathrm{sy}}$ to implement binary synapses as well as multistable synapses with many hundreds of levels between $I_{\mathrm{sy}}^{\mathrm{min}}$ and  $I_{\mathrm{sy}}^{\mathrm{max}}$. To implement unsupervised learning, it is necessary for neuron-generated signals to be capable of modifying the flux in the SS loop. In Secs.\,\ref{sec:Hebbian} and \ref{sec:stdp} we show how photonic signals can be used to change the state of flux in the SS loop and therefore implement learning rules based on timing correlations between the two neurons associated with each synapse.  
	
\section{\label{sec:supervised}Supervised learning}
Several quantities determine the behavior of a synapse. These include the minimum and maximum values of the synaptic weight and the number of increments between the two. For many applications in machine learning, neural networks, and neuroscience, synapses are treated as binary elements that can switch between strong (potentiated) and weak (depressed) states \cite{amfu1994,fuab2007,lide2015}. Memory storage times can be improved if synapses have a large number of stable states between maximally potentiated and depressed states \cite{fuab2007}. In Ref.\,\onlinecite{sh2018b} it was shown that, in a superconducting optoelectronic neuron, changing $I_{\mathrm{sy}}$ from 1\,\textmu A to 3\,\textmu A changes the contribution to the neuron's integrated signal by a factor of 15. In this section, we demonstrate that the synaptic current can be changed over this range by adding anywhere from one to hundreds of fluxons to the SS loop, thereby achieving a range of synapses from a simple binary synapse to a multistable synapse with a pseudocontinuum of stable values.
\begin{figure} 
	\centerline{\includegraphics[width=8.6cm]{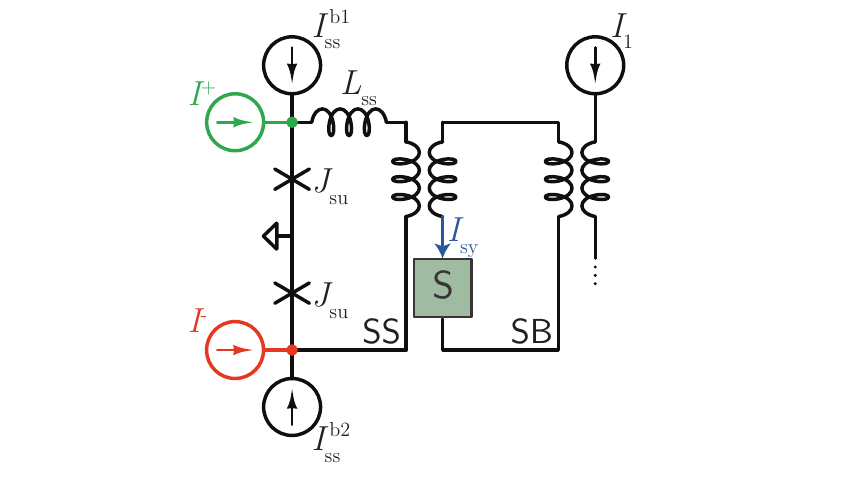}}
	\caption{\label{fig:synapticPlasticity_binaryCircuit}Fluxon memory cell used to achieve binary synapse. The box labeled $\mathsf{S}$ is the synapse receiving the bias current (see Fig.\,\ref{fig:synapticPlasticity_schematic}). Circuit parameters are listed in Appendix \ref{apx:memoryCell}.}
\end{figure}

The circuit for enacting a binary synapse is shown in Fig.\,\ref{fig:synapticPlasticity_binaryCircuit}. This circuit is a standard flux-quantum memory cell \cite{ka1999,vatu1998} coupled to the SB loop via a mutual inductor \cite{miha2005}.  The current delivered to the synapse, $I_{\mathrm{sy}}$, is shifted by a static value determined by the bias $I_1$ and the mutual inductors shown in Fig.\,\ref{fig:synapticPlasticity_binaryCircuit}. When there are no fluxons in the SS loop, $I_{\mathrm{sy}} = 1$\,\textmu A, the minimum value. In this state, the bias currents ($I^{\mathrm{b1}}_{\mathrm{ss}}$ and $I^{\mathrm{b2}}_{\mathrm{ss}}$) are chosen such that a weakening synaptic update signal ($I^-$) cannot add a fluxon to the loop, so the synaptic weight cannot be further depressed. A strengthening signal can, however, switch $J_{\mathrm{su}}$ and add one fluxon to the loop. This transitions the circuit to the potentiated state, wherein $I_{\mathrm{sy}} = 3$\,\textmu A. At this point, further potentiating signals cannot add additional flux to the loop. The loop can store only a single fluxon, and it is characterized by $\beta_{\mathrm{L}}/2\pi = L I_c/\Phi_0 = 1.8$ \cite{ka1999,vatu1998}. The junction and circuit parameters are given in Appendix \ref{apx:memoryCell}. All parameters are typical for superconducting electronic circuits and straightforward to realize in hardware.
\begin{figure} 
	\centerline{\includegraphics[width=8.6cm]{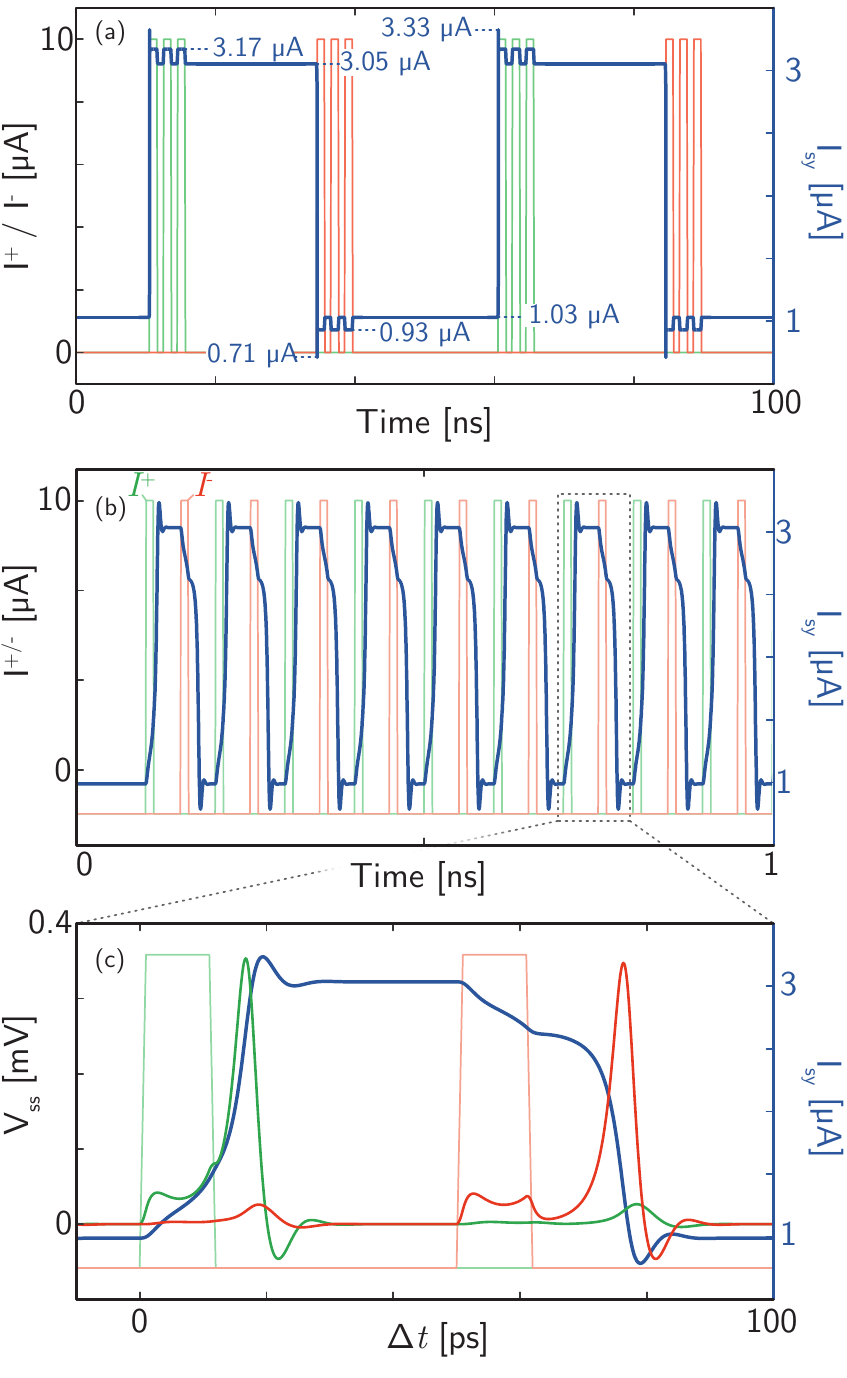}}
	\caption{\label{fig:synapticPlasticity_binary}Operation of binary synapse. (a) Synaptic bias, $I_{\mathrm{sy}}$, as a function of time while potentiating and depressing drive signals are applied. The red and green traces are the drive signals across the two JJs, referenced to the left $y$ axis. The blue trace is $I_{\mathrm{sy}}$, referenced to the right $y$ axis. (b) The operation of the storage loop driven with 50 ps switching time. (c) Temporal zoom of the data in (b).}
\end{figure}

Figure \ref{fig:synapticPlasticity_binary} shows WRSpice \cite{wh1991} simulations of the temporal behavior of the circuit as it switches between states. In Fig.\,\ref{fig:synapticPlasticity_binary}(a), the circuit is initially in the depressed state. A pulse of 10\,\textmu A drives the circuit to the potentiated state. Repeated current pulses do not switch the state, and after the input pulses cease, the cell holds the value of $I_{\mathrm{sy}}$. Upon the application of a single 10\,\textmu A pulse into the weakening port, the circuit switches back to the depressed state, and repeated applications of this signal do not further switch the circuit.

\begin{figure} 
	\centerline{\includegraphics[width=8.6cm]{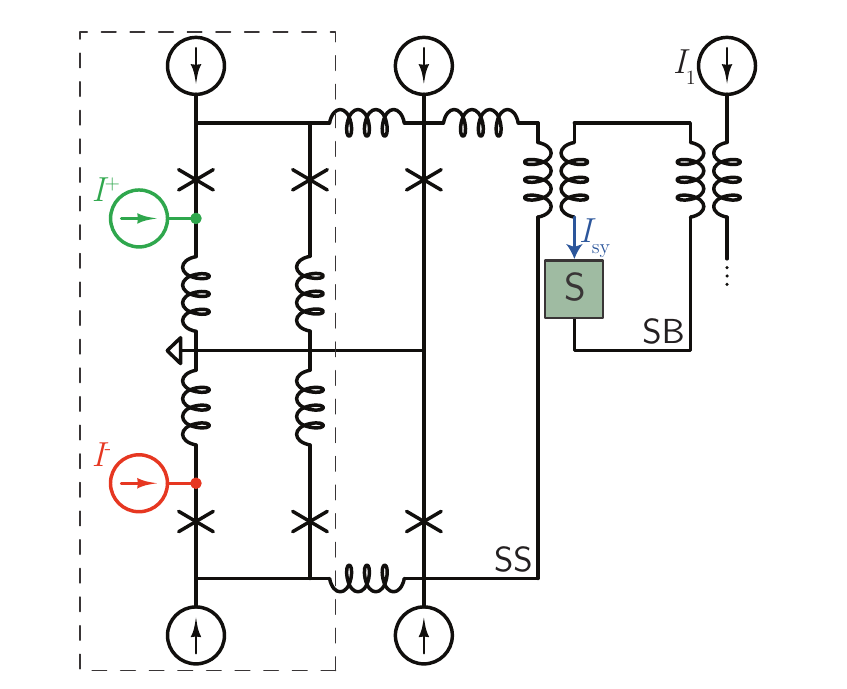}}
	\caption{\label{fig:synapticPlasticity_supervisedCircuit} Diagram of the synaptic update circuit used for supervised learning with multiple stable synaptic values. DC-to-SFQ converters (dashed box) add fluxons to the synaptic storage loop to increase or decrease the synaptic bias current applied to the synaptic firing circuit \cite{sh2018b}. Values of the circuit parameters are listed in Appendix \ref{apx:memoryCell}.}
\end{figure}
In Fig.\,\ref{fig:synapticPlasticity_binary}(b) we show the synapse switching between the depressed and potentiated states every 50\,ps. The time scales of Fig.\,\ref{fig:synapticPlasticity_binary} are extremely fast compared to biological neural circuits. The speed of these circuits offers intriguing possibilities, as discussed in Sec. \ref{sec:discussion}. Figure \ref{fig:synapticPlasticity_binary}(c) shows a temporal zoom of a full cycle of the binary synapse occurring within 50\,ps.    

\begin{figure} 
	\centerline{\includegraphics[width=8.6cm]{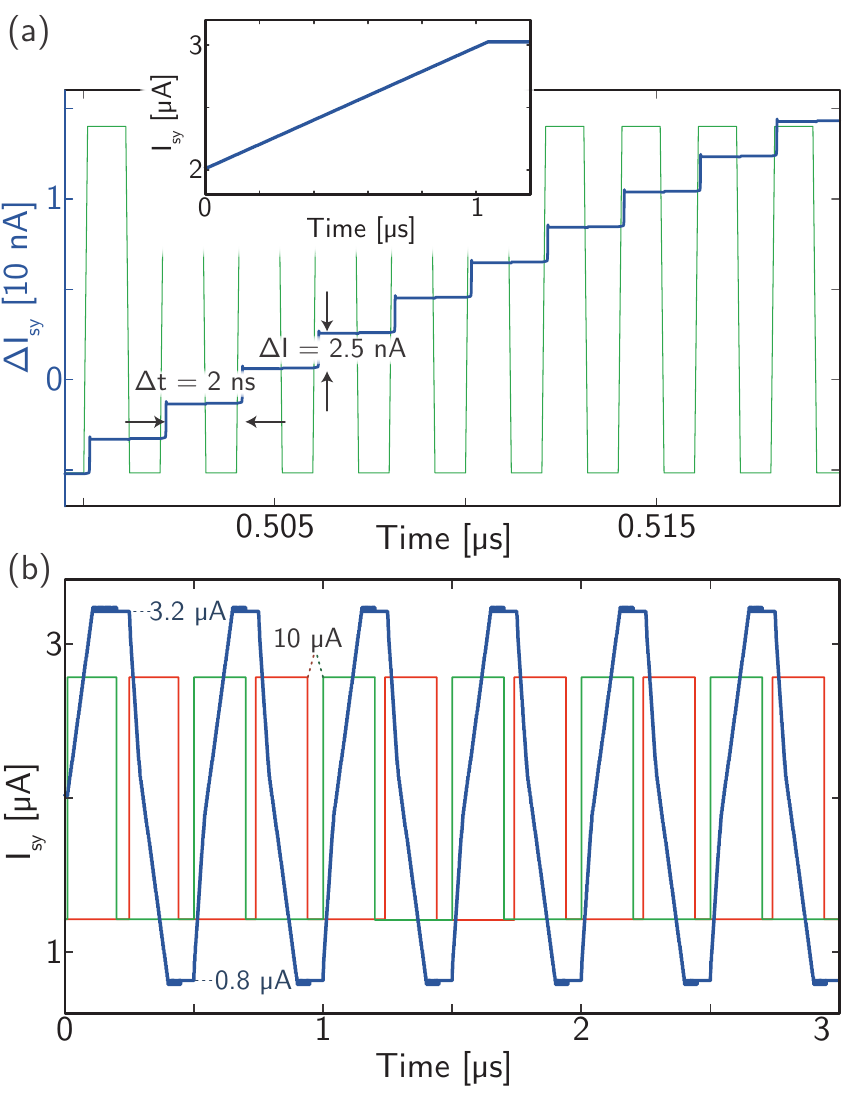}}
	\caption{\label{fig:synapticPlasticity_supervised} Operation of the synaptic update circuit for supervised learning. (a) Synaptic current bias, $I_{\mathrm{sy}}$, is shown by the blue trace, referenced to the left $y$ axis. The drive signal, $I^+$ is shown without reference to an axis. The square wave has a 10\,\textmu A amplitude and 2\,ns period. The main panel shows detail of the operation on a short time scale, and the inset shows a ramp from the middle synaptic weight until saturation at the maximum synaptic weight. In this calculation, $L_{\mathrm{ss}} = 200$\,nH, $\Delta I_{\mathrm{ss}} = 10.3$\,nA per pulse, and $\Delta I_{\mathrm{sy}} = 2.5$\,nA per pulse. (b) Schematic of the circuit used to strengthen as well as weaken synaptic weight. (c) Operation of the circuit as the synaptic weight is repeatedly ramped between minimum and maximum values.}
\end{figure}
For deep learning in neural networks, it is often necessary to increment the synaptic weights in small steps. To achieve fine weight update, a superconducting loop capable of storing more than one flux quantum is utilized, as shown in Fig.\,\ref{fig:synapticPlasticity_supervisedCircuit}. Flux quanta can be added one by one using DC-to-SFQ converters \cite{ka1999,vatu1998}. The binary synapse of Figs.\,\ref{fig:synapticPlasticity_binaryCircuit} and \ref{fig:synapticPlasticity_binary} has been modified to include two DC-to-SFQ converters: one for potentiating and one for depressing.  When a fluxon is produced by the potentiating DC-to-SFQ converter by the introduction of a current pulse, $I^+$, the fluxon is added to the SS loop. When a fluxon is produced by the depressing DC-to-SFQ converter by the introduction of a current pulse, $I^-$, the fluxon counter propagates in the SS loop. The inductors of the SS loop, $L_{\mathrm{ss}}$ and $M_{\mathrm{ss}}$ can be chosen over a broad range of values to determine the learning rate and range of synaptic weights achieved.

Controlled increase of synaptic bias current is again demonstrated using WRSpice \cite{wh1991}. The results are shown in Fig.\,\ref{fig:synapticPlasticity_supervised}. In this calculation, a periodic square wave drives the DC-to-SFQ converter with 10\,\textmu A pulses of 1\,ns duration and 2\,ns period. Current is added to the SS loop in fluxon increments over many input cycles (Fig.\,\ref{fig:synapticPlasticity_supervised}(a)). In this case, the value of $I_{\mathrm{sy}}$ before any flux has been added to the SS loop is 2\,\textmu A, chosen to be in the middle of the operational range identified in Ref.\,\onlinecite{sh2018b}. For this calculation, the inductance of the SS loop is 200\,nH ($\beta_{\mathrm{L}}/2\pi = L I_c/\Phi_0 = 3.8\times 10^3$), leading to the addition of 2.5\,nA to $I_{\mathrm{sy}}$ with the addition of each fluxon to the loop. This value of inductance (and therefore $\Delta I_{\mathrm{sy}}$) can be chosen over a broad range to set the synaptic update increment and number of synaptic levels. This value was chosen to create a SS loop that can store over 1000 fluxons between the minimum and maximum values of $I_{\mathrm{sy}}$. The effects of the number of stable synaptic levels will be discussed further in Sec.\,\ref{sec:discussion}.

The inset of Fig.\,\ref{fig:synapticPlasticity_supervised}(a) shows the behavior of $I_{\mathrm{sy}}$ as a function of time as it is potentiated to saturation. A fluxon is added to the loop every two nanoseconds. After approximately 500 fluxons have been added to the loop, the value of $I_{\mathrm{sy}}$ saturates just above 3\,\textmu A. This saturation behavior is advantageous so that a learning algorithm cannot cause a synaptic weight to grow without bound. 

Figure \ref{fig:synapticPlasticity_supervised}(b) shows $I_{\mathrm{sy}}$ as a function of time as the potentiating and depressing DC-to-SFQ converters are alternately employed, analogous to the two drives of the binary synapse in Fig. \ref{fig:synapticPlasticity_binaryCircuit}. For these calculations, an SS loop with 20\,nH inductance was considered to reduce the time required to achieve saturation. Initially, $I_{\mathrm{sy}} = 2$\,\textmu A. Fluxons are added to the SS loop for 200\,ns, and $I_{\mathrm{sy}}$ reaches its maximum value of 3.2\,\textmu A. Figure \ref{fig:synapticPlasticity_supervised}(b) shows that while the synaptic strengthening drive ($I^+$) is on, once the SS loop reaches saturation, the value of $I_{\mathrm{sy}}$ cannot be increased. The figure further shows that after the synaptic strengthening drive is turned off, $I_{\mathrm{sy}}$ maintains its value (i.e., during the time from 200\,ns - 250\,ns). After 250\,ns, fluxons of the opposite sign begin to be added to the SS loop via the synaptic weakening drive ($I^-$), and $I_{\mathrm{sy}}$ can be driven down to the minimum value (800 nA in this case). Cycling these drives results in the periodic behavior seen in Fig. \ref{fig:synapticPlasticity_supervised}(b). It can be seen that during each strengthening and weakening cycle $I_{\mathrm{sy}}$ versus time has two regions with different slopes. This is due to the fact that when the current in the SS loop is outside a certain range, the DC-to-SFQ converter releases two fluxons per drive cycle. This characteristic is likely of little consequence and may be eliminated with improved circuit design, possibly by separating the DC-to-SFQ converter from the SS loop with a JTL. 

The circuits of Figs.\,\ref{fig:synapticPlasticity_binaryCircuit} and \ref{fig:synapticPlasticity_supervisedCircuit} have several strengths when used to establish the synaptic weight of a superconducting optoelectronic neuron. The nature of the flux-storage Josephson circuits enables cycling and modifying the synaptic weights as many times as necessary without material degradation. The maximum and minimum values of $I_{\mathrm{sy}}$ can be designed to achieve a broad range of operating conditions. Upon reaching the maximum and minimum values, the device saturates, eliminating the possibility of runaway values of synaptic weight. Synaptic update can be carried out in a specified number of increments based on the choice of inductance of the SS loop. The size of these increments will determine the learning/forgetting rate of the synapse. 

While these characteristics of the circuits are conducive to implementing a variety of training algorithms based on back propagation \cite{ni2015} or in conjunction with design through genetic evolution \cite{fldu2008,suma2018}, we would also like to enable systems that learn using only activity within the network. We next consider a Hebbian learning circuit, which strengthens the synaptic weight between two neurons that fire in succession. This will lead to the discussion of a circuit achieving STDP based on the timing correlation between pre- and post-synaptic activity.
	
\section{\label{sec:Hebbian}Hebbian update}
\begin{figure} 
	\centerline{\includegraphics[width=8.6cm]{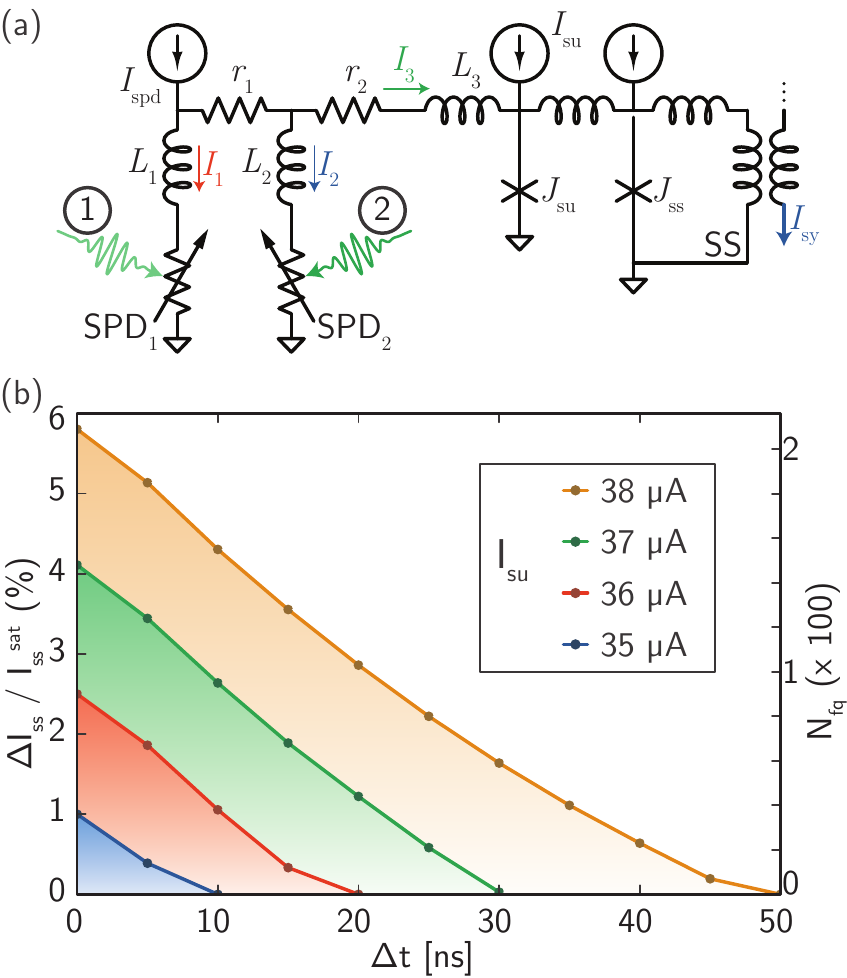}}
	\caption{\label{fig:synapticPlasticity_Hebbian_1}Hebbian update. (a) Hebbian synaptic update circuit diagram. Implements a rule based on temporally correlated single photon detection events from the two neurons associated with the synapse. Circuit parameters are given in Appendix \ref{apx:Hebbian}. (b) Amount of current added to the synaptic storage loop, $\Delta I_{\mathrm{ss}}$, as a percentage of the saturation current of the loop, $I_{\mathrm{ss}}^{\mathrm{sat}}$, versus time delay between upstream and local synaptic update photons, $\Delta t$, for four values of $J_{\mathrm{su}}$ bias current, $I_{\mathrm{su}}$. In these calculations, $L_{\mathrm{ss}} = 1$\,\textmu H.}
\end{figure}
The Hebbian update circuit under consideration is shown in Fig. \ref{fig:synapticPlasticity_Hebbian_1}(a). The operation of this circuit is based on a similar principle to the supervised learning circuits discussed in Sec.\,\ref{sec:supervised} in that the synaptic bias current $I_{\mathrm{sy}}$ is adjusted based on the amount of flux stored in the SS loop. In this section we will explore how the DC-to-SFQ converter of Fig.\,\ref{fig:synapticPlasticity_supervisedCircuit} can be replaced by SPDs to enable flux to be added to the SS loop based on temporally correlated photonic activity within the network. In particular, we wish to implement a Hebbian update rule that potentiates a synaptic connection between pre- and post-synaptic neurons when the pre-synaptic neuron contributes to the firing of the post-synaptic neuron \cite{daab2001}. 

In Ref.\,\onlinecite{sh2018b}, circuits transducing photonic signals to supercurrent are discussed. The Hebbian rule requires a two-photon temporal-correlation circuit, like the temporal-code receiver of Ref.\,\onlinecite{sh2018b}, except the asymmetry of Hebbian update requires an asymmetrical initial bias to the two correlated SNSPDs. Operation of the Hebbian update circuit discussed here can be described qualitatively as follows. When no photons have been detected, the bias $I_{\mathrm{spd}}$ is directed through SPD$_1$. The resistor $r_1$ ensures that SPD$_2$ is unbiased until SPD$_1$ receives a photon, and therefore photons incident on SPD$_2$ have no effect on the circuit unless they are incident during a time window following a detection event by SPD$_1$. Once a photon has been detected by SPD$_1$, $I_{\mathrm{spd}}$ is redirected to $I_2$ and $I_3$. The current returns to $I_1$ with a time constant of $\tau_1 = L_1/r_1$. If a photon is detected by SPD$_2$ during $\tau_1$, $I_{\mathrm{spd}}$ is predominantly redirected to $I_3$, which can be sufficient to switch $J_{\mathrm{su}}$, the synaptic update JJ, perhaps many times depending on the bias currents, $I_{\mathrm{spd}}$ and $I_{\mathrm{su}}$, and the difference in arrival times between the two photons, $\Delta t$. More details of circuit design are included in Appendix \ref{apx:Hebbian}.

During circuit operation, we assume that when the pre-synaptic neuron fires a photonic pulse, one or more photons will reach a synaptic firing circuit \cite{sh2018b} of the post-synaptic neuron and bring the neuron closer to its threshold \cite{sh2018a}. We also assume additional photons have a probability of reaching SPD$_1$ of the synaptic update circuit shown in Fig.\,\ref{fig:synapticPlasticity_Hebbian_1}(a) to perform the first step in implementing the Hebbian rule. This photon is labeled ``1'' in Fig.\,\ref{fig:synapticPlasticity_Hebbian_1}(a). The probability of reaching SPD$_1$ may be controlled to modify the learning rate. Similarly, it is assumed that during a neuronal firing event, the local neuron will send photons to its downstream connections, but also to its local synapse update circuits to activate learning by striking SPD$_2$. This photon is labeled ``2'' in Fig.\,\ref{fig:synapticPlasticity_Hebbian_1}(a). This self-feedback is also illustrated in Fig.\,\ref{fig:synapticPlasticity_schematic}(a).

The duration of the time window after the detection of the pre-synaptic update photon while the circuit is sensitive to the detection of the post-synaptic update photon is determined by the time constant $L_1/r_1$. In Fig.\,\ref{fig:synapticPlasticity_Hebbian_1} we analyze the current added to the SS loop as a function of the delay, $\Delta t$, for four values of $I_{\mathrm{su}}$ with $I_{\mathrm{spd}}$ fixed at $10~\mu$A. We plot the change in current in the SS loop ($\Delta I_{\mathrm{ss}}$) as a percentage of the SS loop saturation current ($I_{\mathrm{ss}}^{\mathrm{sat}}$), during Hebbian update events characterized by delay $\Delta t$. This plot also shows the number of fluxons created during each of the events. We see that the amount of synaptic weight modification depends strongly on the temporal delay, dropping to zero after roughly $\tau_1$. We also see that the effect depends on $I_{\mathrm{su}}$, providing a means by which the memory update rate can be dynamically adjusted during operation via a DC bias current. This dependence on $I_{\mathrm{su}}$ provides a means to implement metaplasticity, as will be discussed in Sec.\,\ref{sec:discussion}. The quantity $\Delta I_{\mathrm{ss}}/I_{\mathrm{ss}}^{\mathrm{sat}}$ represents the fraction of the synapse dynamic range that is acquired in a synaptic update event. Although the current in the SS loop (and therefore $I_{\mathrm{sy}}$) can only change by an integer number of flux quanta, the use of high-kinetic-inductance flux storage loops wherein thousands of flux quanta can be stored makes this effectively an analog circuit. For the SS loop investigated in Fig.\,\ref{fig:synapticPlasticity_Hebbian_1}(a), $\beta_{\mathrm{L}}/2\pi = 1.9\times 10^4$. 

Hebbian learning rules may be based on average firing rates of pre- and post-synaptic neurons or on timing between individual spikes from these neurons \cite{geki2002}. Here we consider the latter. A timing-dependent learning rule often takes the form of exponential decay as a function of the difference in arrival times of pre- and post-synaptic signals. The form shown in Fig.\,\ref{fig:synapticPlasticity_Hebbian_1}(b) is slightly different due to Josephson nonlinearities. This modified temporal dependence is likely of little consequence as it maintains the principal function of timing-dependent plasticity, which is to modify the synaptic weight based on temporal correlation within a specified time window surrounding a neuronal firing event. 

While the quantity $\Delta I_{\mathrm{ss}}$ represents the change in synaptic weight due to one Hebbian update event, the area under the curves in Fig.\,\ref{fig:synapticPlasticity_Hebbian_1}(b) will be related to the learning rate when averaged over many events, because the delay between the two photons, $\Delta t$, will vary across events. In Fig.\,\ref{fig:synapticPlasticity_Hebbian_1}(b), the integral of the curve with $I_{\mathrm{su}} = 35$\,\textmu A is 3.6\% of the integral of the curve with $I_{\mathrm{su}} = 38$\,\textmu A. For $I_{\mathrm{su}} = 36$\,\textmu A, the value is 18\%, and for $I_{\mathrm{su}} = 37$\,\textmu A, the value is 48\%. This indicates we can dynamically change the learning rate across a broad range by adjusting $I_{\mathrm{su}}$. A metaplasticity circuit accomplishing this is discussed in Appendix \ref{apx:metaplasticity}.

\begin{figure} 
	\centerline{\includegraphics[width=8.6cm]{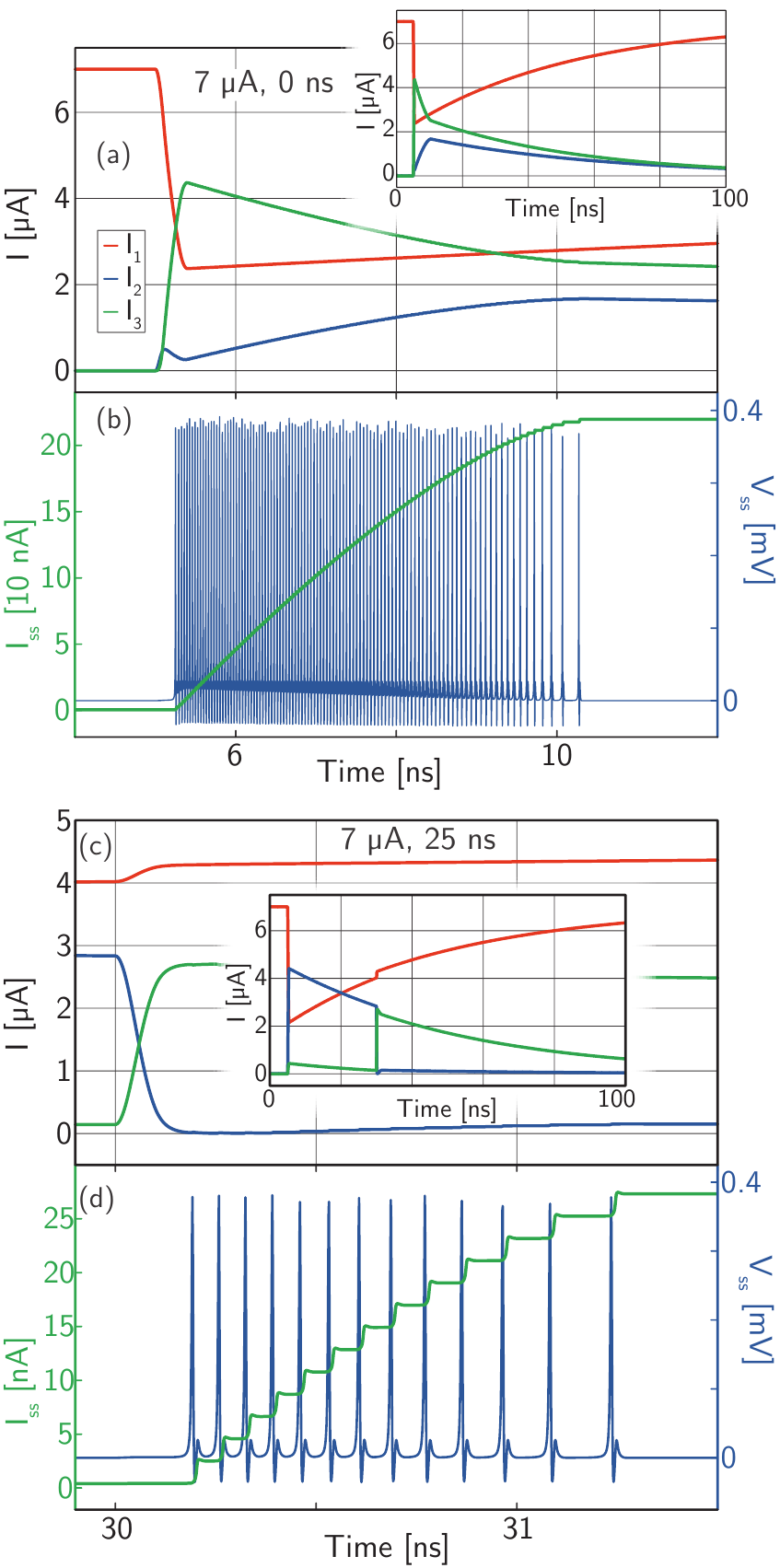}}
	\caption{\label{fig:synapticPlasticity_Hebbian_2}Circuit operation during synaptic update events. (a) Currents $I_1$, $I_2$, and $I_3$ during a synaptic update event with $I_{\mathrm{spd}} = 7$\,\textmu A and $\Delta t = 0$\,ns. The inset shows the currents over a longer time period after photon arrival. (b) The current stored in the synaptic storage loop and the voltage pulses corresponding to 106 fluxons entering the loop. (c) Currents $I_1$, $I_2$, and $I_3$ during synaptic update event with $I_{\mathrm{spd}} = 7$\,\textmu A and $\Delta t = 25$\,ns. (d) The current stored in the synaptic storage loop and the voltage pulses corresponding to 13 fluxons entering the loop.}
\end{figure}

To further illustrate the performance of this device and provide intuition regarding operation, Fig.\,\ref{fig:synapticPlasticity_Hebbian_2} shows details of the operation during Hebbian update events for cases with $\Delta t = 0$\,ns (Fig.\,\ref{fig:synapticPlasticity_Hebbian_2}(a) and (b)) and with $\Delta t = 25$\,ns (Fig.\,\ref{fig:synapticPlasticity_Hebbian_2}(c) and (d)). In these calculations, the SPDs were modeled in WRSpice as transient resistances of 5\,k$\Omega$ lasting for 200\,ps introduced at a specified moment of photon detection. Figures \ref{fig:synapticPlasticity_Hebbian_2}(a) and (c) show the currents $I_1$, $I_2$, and $I_3$ during the time when the junction is switching. The insets of Figs.\,\ref{fig:synapticPlasticity_Hebbian_2}(a) and (c) show a 100\,ns window, capturing the SPD recovery over a longer time. Figures \ref{fig:synapticPlasticity_Hebbian_2}(b) and (d) show the increase in the current circulating in the SS loop as well as the voltage pulses as fluxons enter the loop. We see that the Hebbian update events introduce 25\,nA - 200\,nA to the SS loop. With these values, 280 events with $\Delta t = 25$\,ns or 35 events with $\Delta t = 0$\,ns will saturate the SS loop. The number of events that saturate the loop can be adjusted with the SS loop inductance and with $I_{\mathrm{su}}$.

While it is helpful to demonstrate a Hebbian update mechanism using two photons coupled to a simple JJ circuit, learning rules that can both strengthen and weaken the synaptic connection are required for neural computing.
	
\section{\label{sec:stdp}Spike-timing-dependent plasticity}
\begin{figure} 
	\centerline{\includegraphics[width=8.6cm]{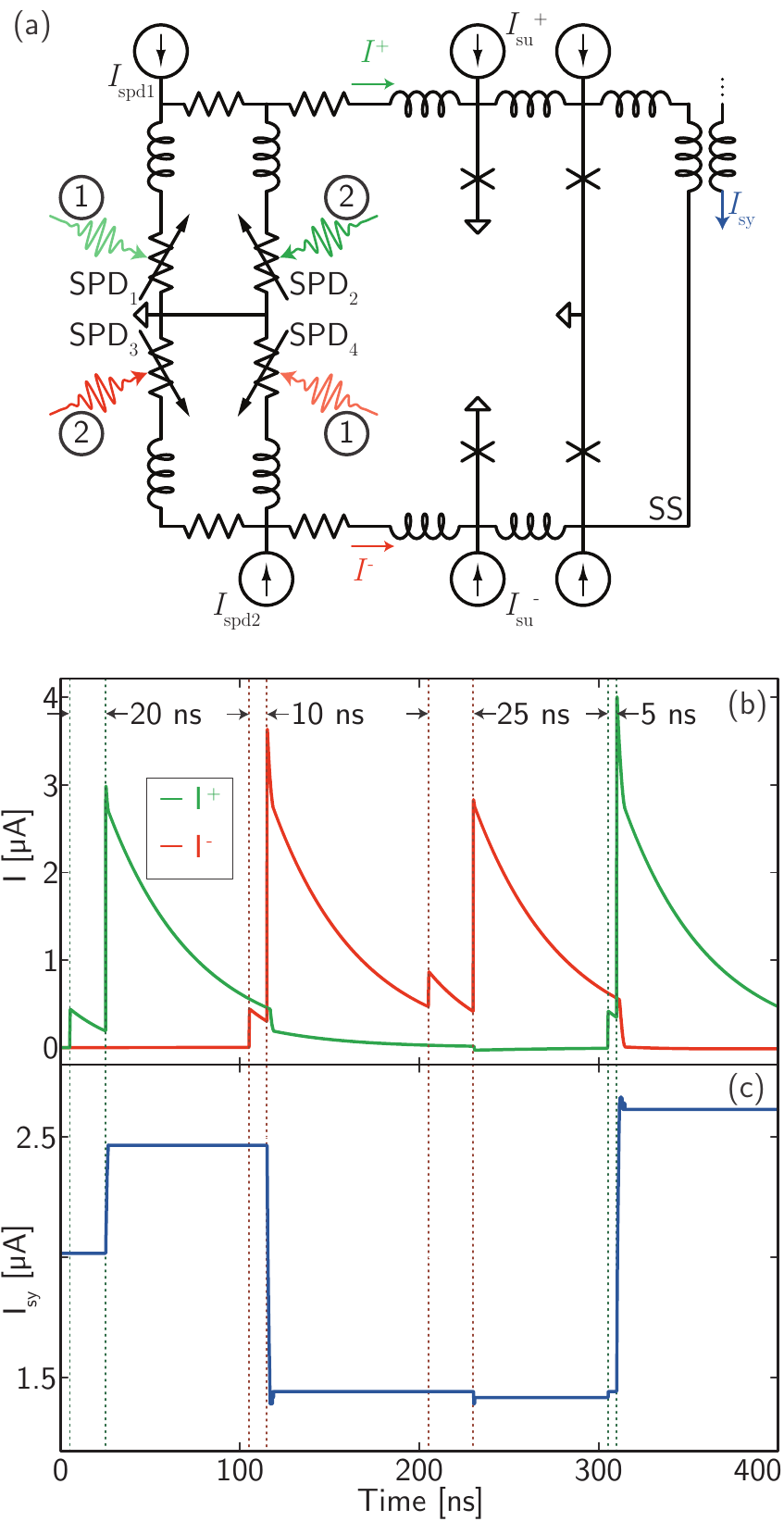}}
	\caption{\label{fig:synapticPlasticity_stdp}Implementation of spike-timing-dependent plasticity. (a) Circuit under consideration. (b) The currents $I^+$ and $I^-$ during synaptic update events. (c) The synaptic bias current, $I_{\mathrm{sy}}$, delivered to the synaptic firing circuit \cite{sh2018b} as a function of time as the synaptic update events of (b) strengthen and weaken the synaptic connection. The full circuit with the STDP module (a) delivering $I_{\mathrm{sy}}$ to the synaptic firing junction is shown in Fig.\,\ref{fig:synapticPlasticity_fullCircuit} of Appendix \ref{apx:fullCircuit}.}
\end{figure}

The STDP we seek to implement performs the Hebbian potentiating operation described in Sec.\,\ref{sec:Hebbian}, but also enforces an anti-Hebbian depressing rule wherein a neuronal firing event at the post-synaptic neuron followed closely by a neuronal firing event at a pre-synaptic neuron depresses the synaptic weight between the two neurons. A circuit capable of producing this STDP is depicted in Fig.\,\ref{fig:synapticPlasticity_stdp}(a). Much as strengthening and weakening were accomplished in Sec.\,\ref{sec:supervised} by adding a mirror image of the strengthening circuit to the SS loop, here we duplicate the Hebbian circuit of Sec.\,\ref{sec:Hebbian} to achieve STDP. The similarity of the SPD circuit of Fig.\,\ref{fig:synapticPlasticity_stdp}(a) and the JJ circuit of Fig.\,\ref{fig:synapticPlasticity_supervised}(a) is apparent. 

The symmetry between the strengthening and weakening receiver circuits in the STDP circuit of Fig.\,\ref{fig:synapticPlasticity_stdp}(a) is broken based on whether the SPD that is biased in the steady state receives photons from the pre-synaptic or post-synaptic neuron. In the synaptic-weakening receiver circuit, a post-synaptic photon detected by SPD$_3$ followed by a photon from a pre-synaptic neuron detected by SPD$_4$ introduces counter-circulating flux to the SS loop. The time constants and biases of the strengthening and weakening receivers can be adjusted independently. 

The WRSpice calculation shown in Fig.\,\ref{fig:synapticPlasticity_stdp}(b) and (c) illustrates the circuit in operation. Two synaptic strengthening events and two synaptic weakening events occur. The currents associated with synaptic strengthening and weakening, $I^+$ and $I^-$ are shown in Fig.\,\ref{fig:synapticPlasticity_stdp}(b). The synaptic bias current delivered to the synaptic firing junction \cite{sh2018b} is shown in Fig.\,\ref{fig:synapticPlasticity_stdp}(c). 
A synaptic strengthening event occurs with $\Delta t = 20$\,ns, followed by a weakening event with $\Delta t = 10$\,ns and another with $\Delta t = 25$\,ns. A final strengthening event occurs with $\Delta t = 5$\,ns. The synaptic bias current, $I_{\mathrm{sy}}$, is observed to respond as expected based on the Hebbian analysis in Sec.\,\ref{sec:Hebbian}. In this calculation, $L_{\mathrm{ss}} = 20$ nH, and we mention again that the amount of current added to $I_{\mathrm{ss}}$ and therefore $I_{\mathrm{sy}}$ during a synaptic update event can be linearly scaled with $L_{\mathrm{ss}}$ in hardware and with $I_{\mathrm{spd}}$ dynamically. Below a certain value of $I_{\mathrm{spd}}$ photon detection will not occur, and memory update will cease. The memory update rate of the STDP synapse can be controlled by adjusting the frequency of photon absorption events. This consideration and others related to implementation of these circuits are discussed in Sec.\,\ref{sec:discussion}. 

The circuit of Fig.\,\ref{fig:synapticPlasticity_stdp}(a) induces STDP based on photon detection events from the pre- and post-synaptic neurons. It may also be possible to achieve STDP entirely in the electronic domain through the electrical signals produced during synaptic firing events and neuronal firing events. One means to use fluxon signals while setting correlation time constants with $L/r$ (as shown in Fig.\,\ref{fig:synapticPlasticity_stdp}) is to use fluxons to switch the gate of an nTron \cite{mcbe2014}. In Ref.\,\onlinecite{sh2018d} we show this operation in the context of a neuronal thresholding element. For STDP, the SPDs could be replaced with nTrons. Fluxons generated by $J_{\mathrm{sf}}$ during synaptic firing events would represent pre-synaptic activity and would switch the gates of nTrons replacing the left SPDs in Fig.\,\ref{fig:synapticPlasticity_stdp}(a). Fluxons generated by the thresholding junction, $J_{\mathrm{th}}$ \cite{sh2018d}, would switch the gates of nTrons replacing the right SPDs in Fig.\,\ref{fig:synapticPlasticity_stdp}(a). Hebbian and anti-Hebbian rules would be implemented based on temporal correlation between pre- and post-synaptic activity, and no photons would need to be expended for the operation. Yet the complexity of Josephson circuitry at each synapse would increase. 

While crucial to learning and the interplay between the structure and function of neural systems, STDP is only one of many synaptic plasticity mechanisms. Despite their significance, discussion of short-term plasticity, homeostatic plasticity, and metaplasticity are relegated to Appendices \ref{apx:shortTerm}, \ref{apx:homeostatic}, and \ref{apx:metaplasticity}. Discussion of how the parameters of the circuits presented here map to learning rate and enable quick adaption alongside long-term memory retention is presented in Appendix \ref{apx:learningRate}.
	
\section{\label{sec:discussion}Discussion}
This work has explored synaptic update circuits capable of delivering a variable synaptic bias current to the synaptic firing circuits presented in Ref. \onlinecite{sh2018b}. We have investigated manipulation of the synaptic weight through external input of square wave pulses, as would be desirable for supervised learning, as well as manipulation of synaptic weight via photon detection events, as would be desirable for unsupervised learning. As an extension of supervised learning, it is interesting to consider using JJ circuits for fast control of synaptic weights. Using Josephson driver circuits \cite{hehe2011,li2012,bewa2015}, synaptic weights could be precisely dynamically controlled. In Sec.\,\ref{sec:supervised} we showed simulations of cycling between weak and strong synaptic weights with no perceptible hysteresis at 10\,GHz. Firing rates of superconducting optoelectronic neurons are likely to be limited to below 1\,GHz due to SPD recovery time and emitter lifetime. Operation with network cycles having oscillation frequencies up to 20 MHz is likely. The potential to vary synaptic weight at much higher frequencies (10\,GHz) introduces the possibility that synaptic connections could be weighted in the frequency domain. The same synaptic weight between two neurons could be strong in some Fourier components and weak in others.

For unsupervised learning, we considered circuits combining single\textendash photon detectors and Josephson junctions to implement unsupervised synaptic update rules based on photons received from correlated neuronal firing events. For full spike\textendash timing\textendash dependent plasticity, the synaptic update circuits described here provide ports for four photons: one strengthening photon from both the pre\textendash synaptic and post\textendash synaptic neuron, and one weakening photon from both the pre\textendash synaptic and post\textendash synaptic neuron. For a single synaptic strengthening or weakening event, two of these photons must be present. When optically implementing a synaptic update rule based on timing correlation, it is difficult to achieve a circuit requiring fewer than two photons.

Other forms of photonic synapses have recently been developed and offer utility in multiple neural contexts \cite{prsh2017,tana20142,tafe2017,shha2016,chri2017}. One can leverage phase shifts in microrings \cite{tana20142,tafe2017} or Mach-Zehnder interferometers (MZIs) \cite{shha2016} to adjust synaptic weight. Thermal tuning is often employed to implement the phase shifts. Thermal tuning requires more power than is suitable for this hardware platform. Phase shifters may be also be large if MZIs are used, and phase shifters may require exotic materials, which limit scaling if electro-optic effects are leveraged. If different synaptic channels are addressed with different frequencies of light, the out-degree of a node in the network is limited by the multiplexed channel spacing. Approaches using MZIs for weighting and routing have the disadvantage that STDP cannot be implemented because modifying a single phase shifter in the network affects many synaptic weights. One approach to synaptic weighting in the photonic domain utilizes a variable optical attenuator at each synaptic connection. Phase-change materials have been employed as such variable attenuators \cite{chri2017}, and the absorption of phase change materials can be affected with pulses of light, thus introducing a Hebbian-type synaptic weight update process. While such an approach may be useful for certain types of neural circuits, update of these synapses requires too many photons to be useful for the energy-efficient neural computing scheme developed here (billions of photons per update operation for phase change versus single photons for superconducting optoelectronics). It is also not clear how anti-Hebbian synaptic update can be introduced to enable full spike-timing-dependent plasticity. It remains to be seen if other synaptic operations such as short-term plasticity, homeostatic plasticity, and metaplasticity can be achieved with phase-change materials. Synaptic weights that attenuate a signal in the optical domain require more light from neuronal firing events, and many photons are simply absorbed at weak synapses. By contrast, using photons for communication but weighting in the superconducting domain, as presented here, uses fluxons to change the synaptic weight, and they can be generated with orders of magnitude less energy than photons. While all of these approaches to synaptic weighting may be useful in different contexts, we have developed the synapses presented in this work based on simultaneous considerations of power, complexity, scalability, speed, and size in the context of the superconducting optoelectronic hardware platform \cite{shbu2017,sh2018a}.

An important weakness of the synapses presented here is they lose all memory when superconductivity is broken. The neuromorphic system must remain below $T_c$ to preserve what has been learned. This class of Frosty the Snowman memory may be augmented by devices that can be heated, such as magnetic Josephson junctions \cite{ru2016,scdo2017,scdo2018}. It would be appealing if the state of memory in the plastic synapses described here could be transferred to long-term magnetic memory, perhaps during a sleep phase.

Another potential challenge for this type of memory in loop neurons is flux trapping. The synaptic integration loops discussed in Ref.\,\onlinecite{sh2018b} are likely to include resistors to give a leak rate. Trapped flux in those loops will be less problematic. The synaptic storage loops that set the synaptic weights are intended to store flux for a long time to maintain memory, so they will not include resistors. In this case, trapped flux will produce variations in the initial synaptic weights across an ensemble. For binary synapses, this will result in some synapses being initialized with strong synaptic weight, and some with weak. For SS loops with high inductance, stray flux will induce a small current, so the perturbation may be small relative to the dynamic range of the synapse. For large ensembles of synapses, the statistical variation may be tolerable or even advantageous. If flux proves problematic, techniques used to shield superconducting qubits can be employed \cite{coch2011}.

In Ref.\,\onlinecite{sh2018a} we argue that a dynamical system capable of differentiated processing and information integration across spatial and temporal scales underlies cognition. In Ref.\,\onlinecite{sh2018b} we introduced the relaxation oscillators and dendritic processing loops capable of implementing the temporal synchronization operations necessary for integrating information in time. Network synchronization and synaptic plasticity are mutually constructive phenomena in that synaptic strengthening through spike timing is more likely to occur when the firing of two neurons is correlated, and the strengthened synapses, in turn, make the correlated neurons more likely to synchronize. Networks with small-world structure \cite{wast1998,sp2010} and dynamics characterized by self-organized criticality are crucial to achieving information integration. Hebbian learning rules and STDP have also been shown to convert random networks into small-world networks and to give rise to self-organized criticality \cite{siqu2007,rusp2011}. Creation of hardware capable of supporting complex networks and synaptic learning mechanisms will provide a powerful tool for the investigation of the relation between critical network dynamics and cognitive function. In the present work we have shown the complex synaptic behavior necessary for rapid adaptation, long-term memory retention, and synaptic update based on network activity. Networks of neurons connected by these synapses will be capable of integrating information learned at many times in many contexts in a single dynamical state.

This work has focused on changing synaptic weights in superconducting optoelectronic neurons. A central question of the hardware platform remains: how are the photons created? This question is addressed in the next paper in this series, Ref.\,\onlinecite{sh2018d}.

\vspace{0.5em}
This is a contribution of NIST, an agency of the US government, not subject to copyright.
	
\newpage
\appendix
	
\section{\label{apx:memoryCell}Circuit parameters of supervised memory cells}
The memory cell of Fig. \ref{fig:synapticPlasticity_binaryCircuit} has been designed with the following circuit parameters. $I_{\mathrm{ss}}^{\mathrm{b1}} = 38$\,\textmu A, $I_{\mathrm{ss}}^{\mathrm{b2}} = 20$\,\textmu A, $L_{\mathrm{ss}} = 90$\,pH. The four inductors comprising the two mutual inductors are labeled $L_1-L_4$ from left to right. Their values are $L_1=L_2=45$\,pH, $L_3=L_4=18$\,pH.

The memory cell of Fig. \ref{fig:synapticPlasticity_supervisedCircuit} has been designed with the following circuit parameters. The inductors comprising the DC-to-SFQ converter are, from left to right, $L_1 = 80$\,pH, $L_2 = 60$\,pH, $L_3 = 300$\,pH. The bias to the DC-to-SFQ converter is $I_{\mathrm{DC}} = 73$\,\textmu A. The drive current pulses are $I_{+} = 10$\,\textmu A with 100\,ps rise and fall time and 1 ns duration. The bias to the JJ in the SS loop is $I_{\mathrm{ss}}^{\mathrm{b}} = 34$\,\textmu A. The mutual inductor parameters between the SS loop and the SB loop and from the SB loop to $I_1$ are, from left to right $L_1 = 18$\,pH, $L_2 = 190$\,pH, $L_3 = 18$\,pH, $L_4 = 18$\,pH, and $I_1 = 27$\,\textmu A. With $L_{\mathrm{ss}} = 20$\,nH, $\Delta I_{\mathrm{ss}} = 103$\,nA per pulse, $\Delta I_{\mathrm{sy}} = 25$\,nA per pulse. The SS loop can store $-4.94$\,\textmu A $< I_{\mathrm{ss}} < 4.96$\,\textmu A.

In this work, all Josephson junctions have $I_c = 40$\,\textmu A. In contrast to the circuits of Ref.\,\onlinecite{sh2018b} where JJs with $I_c = 10$\,\textmu A were used, these JJs do not switch with every synaptic firing event, and consequently, using lower $I_c$ for power minimization is less important. Using $I_c = 40$\,\textmu A leads to circuits with wider operating margins and ease of fabrication. We argue in Refs.\,\onlinecite{sh2018d} and \onlinecite{sh2018e} that using JJs with $I_c = 40$\,\textmu A for the circuits of Ref.\,\onlinecite{sh2018b} would also be satisfactory. The JJs in this work \cite{sh2018b} have been simulated with $\beta_c = 0.95$, corresponding to slightly over-damped junctions \cite{vatu1998,ka1999}.
	
\section{\label{apx:Hebbian}Considerations for Hebbian circuit design}
To achieve the desired Hebbian operation with the circuit of Fig.\,\ref{fig:synapticPlasticity_Hebbian_1}(a), several considerations are pertinent. When SPD$_1$ detects a photon, it needs to direct current predominantly to $I_2$, and not to $I_3$. When SPD$_2$ detects a photon, it needs to direct current predominantly to $I_3$, and not to $I_1$. These considerations inform us that we should choose $L_2 \ll L_3$ and $L_3 \ll L_1$. We choose $L_2 = 12.5$\,nH for this study. Such a small SPD may have reduced detection efficiency, but the inefficiency is tolerable for this purpose, because synaptic update will occur only rarely to optimize memory retention \cite{fuab2007,lide2015}. We then choose $L_3 = 125$ nH, and $L_1 = 1.25$\,\textmu H. The choices for $r_1$ and $r_2$ are made to achieve the desired temporal behavior. The $L/r$ time constants must be long enough to ensure the SPDs do not latch. Beyond this, they can be chosen to achieve the desired learning performance. We choose $\tau_1 = 50$ ns and $\tau_2 = 5$\,ns to facilitate WRSpice analysis, but longer time constants may be necessary in practice.

The circuit parameters relevant to Fig.\,\ref{fig:synapticPlasticity_Hebbian_1}(a) are as follows. Inductor values are $L_1 = 1.25$\,\textmu H, $L_2 = 12.5$\,nH, $L_3 = 125$\,nH. $I_{\mathrm{spd}} = 7$\,\textmu A - 10\,\textmu A. The bias to the synaptic update junction is $I_{\mathrm{su}}^{\mathrm{b}} = 38$\,\textmu A, and the bias to the synaptic storage junction is the same. The resistors $r_1$ and $r_2$ can be chosen to achieve the desired correlation time window.
	
\section{\label{apx:fullCircuit}Synaptic update circuit supplying synaptic firing circuit}
\begin{figure} 
	\centerline{\includegraphics[width=8.6cm]{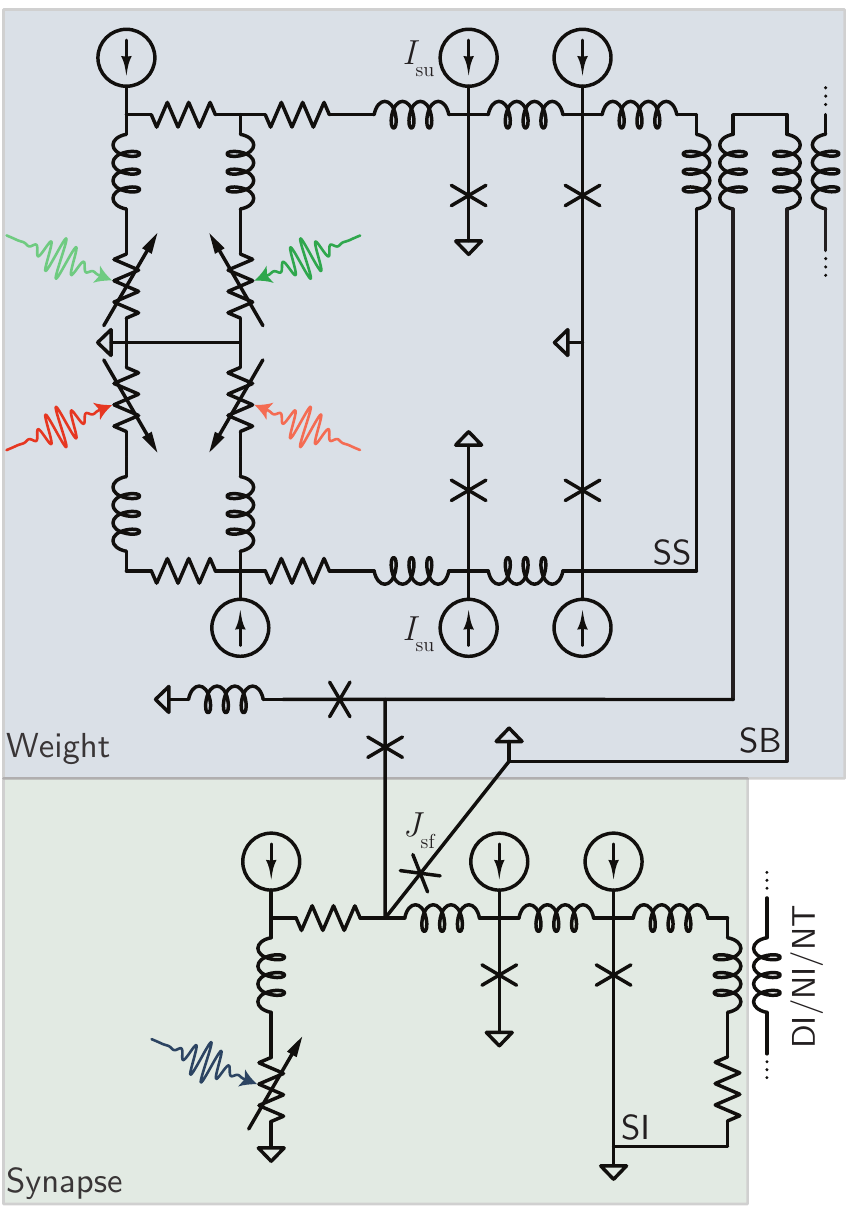}}
	\caption{\label{fig:synapticPlasticity_fullCircuit}Synaptic update circuit supplying $I_{\mathrm{sy}}$ to synaptic firing circuit.}
\end{figure}

The circuit configuration combining the synaptic update circuit of this work with the synaptic firing circuit of Ref.\,\onlinecite{sh2018b} is shown in Fig.\,\ref{fig:synapticPlasticity_fullCircuit}. Bias current $I_1$ can be used to supply many synapses. A buffer stage ($J_{\mathrm{b1}}$ and $J_{\mathrm{b2}}$) isolates the SI loop from flux generated during synaptic firing events.
Initial simulations of this configuration show that the buffer can employ junctions with $J_{\mathrm{b1}}$ having 10\,\textmu A $I_c$ (same as $J_{\mathrm{sf}}$ of Ref.\,\onlinecite{sh2018b}), and $J_{\mathrm{b2}}$ with 40\,\textmu A $I_c$ used throughout this work.

\section{\label{apx:shortTerm}Short-term plasticity}
\begin{figure} 
	\centerline{\includegraphics[width=8.6cm]{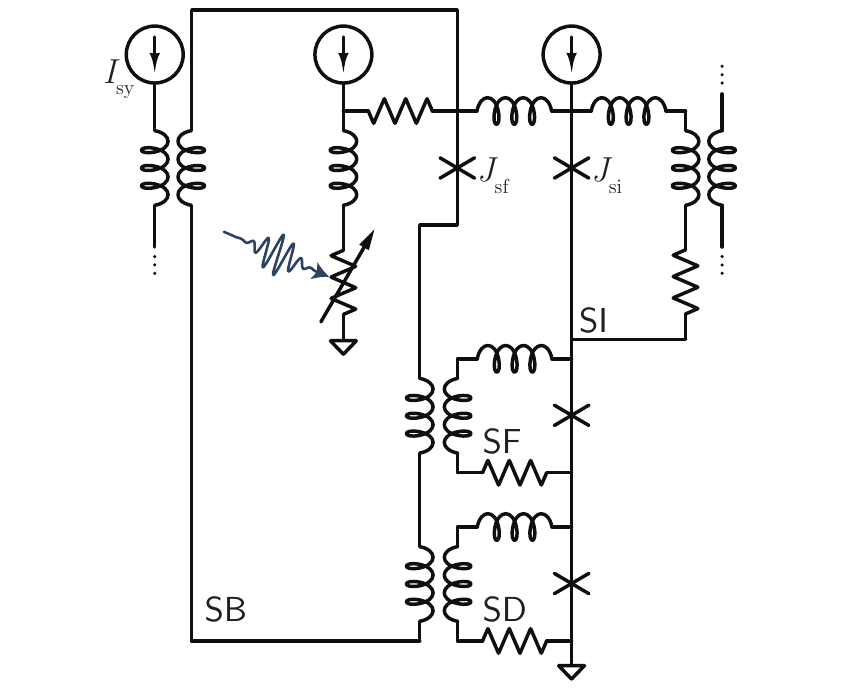}}
	\caption{\label{fig:synapticPlasticity_shortTermCircuit}Circuit for implementing short-term plasticity. Synaptic firing events cause the junctions in the short-term facilitating and short-term depressing loops to generate flux. The flux in the facilitating loop acts to temporarily achieve synaptic gain, and the flux in the depressing loop acts to temporarily suppress the synaptic efficacy. The inductances of the loops determine the magnitude of their effects, and the $L/r$ time constants determine the temporal envelope. Independent control of $L$ and $r$ in each of the loops shapes the filter response.}
\end{figure}

Short term plasticity varies the post-synaptic response to a pre-synaptic pulse train \cite{abre2004} on a time scale close to the inter-spike interval \cite{daab2001}. Short term plasticity acts as a filter, and the response can be low-pass, high-pass, or band-pass depending on a number of factors. These various filtering operations can be achieved in loop neurons with the addition of typical SPD/JJ loop circuits that change their state in response to pre-synaptic activity, either from photons from the pre-synaptic neuron or fluxons generated during the synaptic firing event. A circuit that may be utilized to perform the filtering operations of short-term plasticity using additional JJs and the fluxons produced during a synaptic firing event is shown in Fig.\,\ref{fig:synapticPlasticity_shortTermCircuit}. This circuit expands upon the synaptic receiver circuit of Ref.\,\onlinecite{sh2018b}. Two additional JJs have been added in series to the junction in the Josephson transmission line, $J_{\mathrm{jtl}}$. These JJs are coupled to independent loops that are inductively coupled to the synaptic bias loop. In the absence of synaptic activity, the bias is set by $I_{\mathrm{sy}}$, just as before. However, during a synaptic firing event, the two additional junctions also switch. Therefore, flux is coupled to the synaptic integrating loop, as before, but flux is also added to two new loops, the short-term facilitating loop (SF), and the short-term depressing loop (SD). The SF loop will add current to $I_{\mathrm{sy}}$, effectively strengthening the synaptic weight, and the SD loop will reduce the current to $I_{\mathrm{sy}}$, effectively weakening the synaptic weight. Therefore, the sign of the mutual inductance of the two loops is opposite, and their magnitude may differ. The time constants of the SF and SD loops can be set independently, and they will likely be slightly longer than the inter-spike interval of pulse trains in the system ($\approx$\,1\,\textmu s). A given synapse may employ one or both loops as required for information processing, and a given neuron is likely to benefit from an ensemble of synapses with diverse short-term filtering responses.

Short-term facilitation may operate such that the first pre-synaptic pulse evokes no post-synaptic response, and only after several pulses has the synaptic weight been facilitated to the point of communicating subsequent pulses. The circuit of Fig.\,\ref{fig:synapticPlasticity_shortTermCircuit} would need to be modified to achieve this behavior, as facilitation and depression both depend on the switching of the junctions, which requires successful pre-synaptic transmission. Such short-term facilitating behavior can be accomplished by introducing an additional SPD explicitly for short-term plasticity. This SPD would receive photons from the pre-synaptic neuron, just as the SPD in the receiver circuit shown in Fig.\,\ref{fig:synapticPlasticity_shortTermCircuit}, but the additional SPD would add no flux to the SI loop upon firing, and would instead only adjust the flux in the SF and SD loops.

\section{\label{apx:homeostatic}Homeostatic plasticity}
\begin{figure} 
	\centerline{\includegraphics[width=8.6cm]{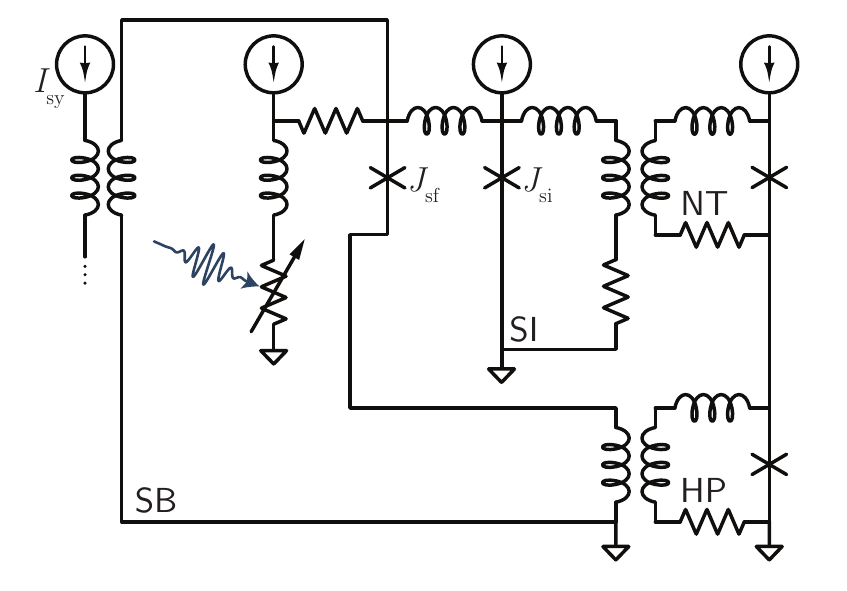}}
	\caption{\label{fig:synapticPlasticity_homeostaticCircuit}Circuit for implementing homeostatic plasticity. A fluxon generated by the thresholding junction during a neuronal firing event changes the flux state of the homeostatic plasticity loop. This flux is inductively coupled to the bias of the synaptic firing junction.}
\end{figure}

The function of homeostatic plasticity is to modulate synaptic efficacy in response to a running average of post-synaptic neuronal activity to keep neuronal gain within a useful dynamic range \cite{cobe2012}. A loop neuron circuit can implement homeostatic plasticity with fluxons generated by thresholding events of the post-synaptic neuron. One means to achieve this operation is shown in Fig.\,\ref{fig:synapticPlasticity_homeostaticCircuit}.

Whereas the short-term plasticity circuit of Appendix \ref{apx:shortTerm} made use only of signals generated by a synaptic firing event, the homeostatic plasticity circuit makes use only of signals generated by post-synaptic neuronal firing events. The homeostatic plasticity (HP) loop is negatively inductively coupled to the synaptic firing junction ($J_{\mathrm{sf}}$), meaning flux added to HP reduces the bias to $J_{\mathrm{sf}}$, thereby depressing the synaptic efficacy. By choosing an $L/r$ time constant for the HP loop that is longer than the neuron's typical inter-spike interval, the negative HP feedback depends on a sliding temporal average of post-synaptic neuronal firing activity \cite{bico1982}.

\section{\label{apx:metaplasticity}Metaplasticity}
Homeostatic plasticity (Appendix \ref{apx:homeostatic}) is one example of a plasticity mechanism that compensates for neural activity on longer time scales to adjust learning rate. Homeostatic plasticity is a response to a sliding temporal average of the post-synaptic neuron \cite{bico1982,cube2012}. Metaplasticity refers more generally to mechanisms that adjust not the synaptic efficacy, but the rate of change (or probability of change) of synaptic efficacy. Here we discuss a loop circuit that achieves a metaplastic response \cite{fudr2005,ab2008,khso2017} based on both pre-synaptic and post-synaptic activity using similar SPD/JJ circuits to those developed for the STDP circuit of Sec.\,\ref{sec:stdp}. 

\begin{figure*} 
	\centerline{\includegraphics[width=17.2cm]{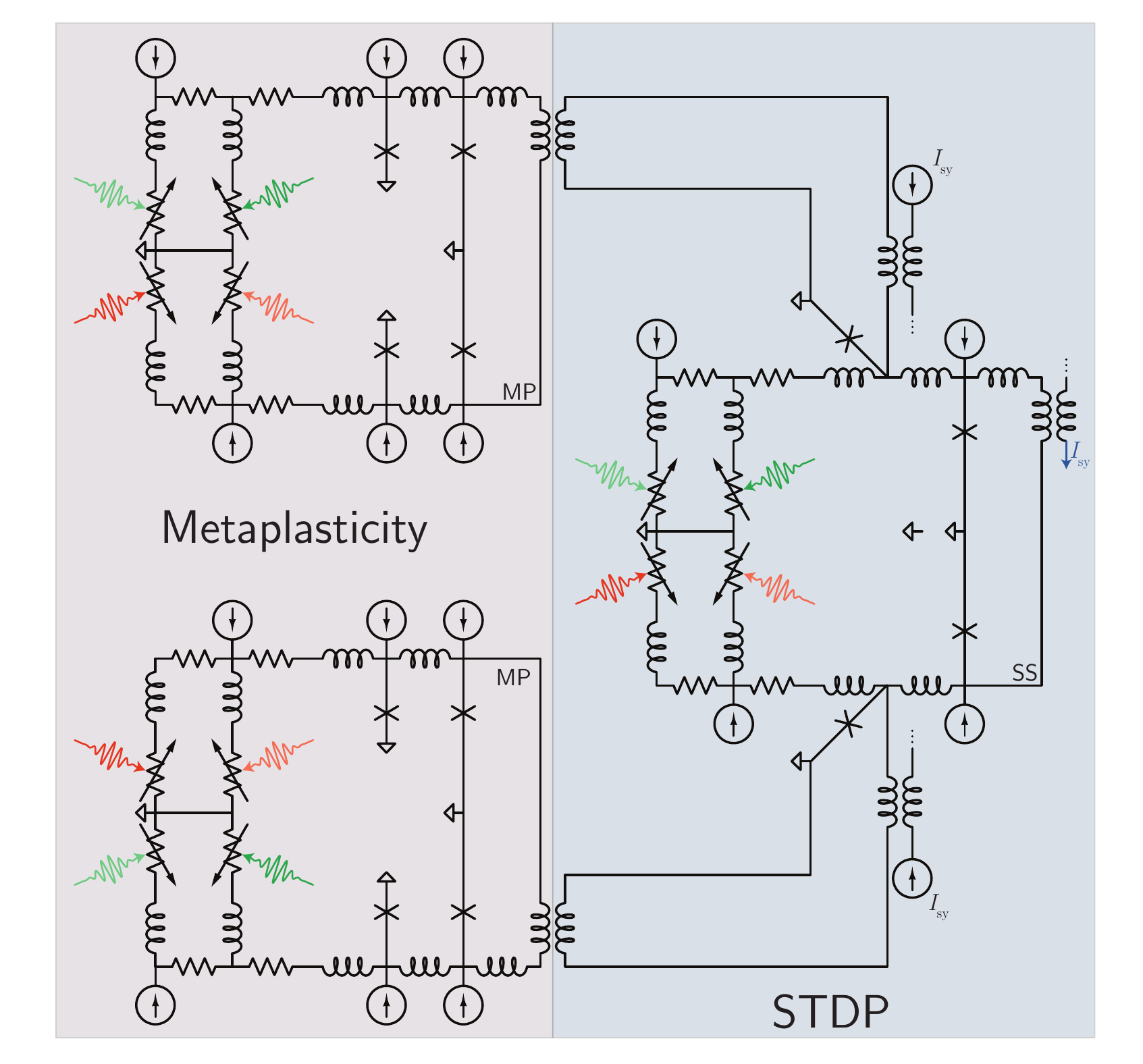}}
	\caption{\label{fig:synapticPlasticity_metaplasticityCircuit}Circuit for achieving metaplasticity with single-photon update events. The metaplasticity loop stores a history of pre-synaptic/post-synaptic correlation events, and this trapped flux is inductively coupled to the bias of the STDP update circuitry. Hebbian/anti-Hebbian correlation circuits are shown, but symmetrical correlation circuits \cite{sh2018b} could also be used. Single-photon detectors could be replaced by nTrons to utilize only local electrical signals.}
\end{figure*}
The circuit under consideration is shown in Fig.\,\ref{fig:synapticPlasticity_metaplasticityCircuit}. The concept here is similar to many other operations in loop neurons. The fractional change in synaptic efficacy ($\alpha$, see Sec.\,\ref{sec:discussion}) incurred during an STDP update event in the STDP circuit of Sec.\,\ref{sec:stdp} depends on the magnitude of the current through the junctions in parallel with the SPDs. The function of the metaplastic circuit of Fig.\,\ref{fig:synapticPlasticity_metaplasticityCircuit} is to modify these bias currents based on correlated pre-synaptic and post-synaptic activity. To achieve this operation, the same circuit block that is employed to adjust $I_{\mathrm{sy}}$ during a plasticity (efficacy update) event is also employed to adjust the JJ bias during a metaplasticity (learning rate update) event. With this circuit, the amount the synaptic efficacy is adjusted during an STDP update event depends on the flux trapped in the metaplasticity (MP) loops. The learning rate depends on both the efficacy update frequency as well as the magnitude of each update (see Sec.\,\ref{sec:discussion}). Thus, by changing the magnitude of the updates, the metaplastic circuits modify the learning rate. 
		
Considering the metaplastic circuits in the context of the synapse as a whole \cite{sh2018b}, the state of the synapse is associated with the flux in the synaptic integrating loop. The rate of change of the state of the synapse is associated with the flux in the synaptic storage loop. The rate of change of the rate of change of the state of the synapse is associated with the flux in the metaplasticity loops. By cascading additional loops, one can continue the hierarchy of synaptic loops that record the state of the synapse and its derivatives. We suspect the three levels of hierarchy presented here will suffice for many applications.

At this point, a basic algorithm for loop neuron design has emerged. For each synaptic function, add an SPD, a JJ, and a loop. Inductively couple the loop bias currents as functionally appropriate. Choose time constants carefully. Repeat until there is no more space. As the number of plasticity operations, and therefore SPDs, JJs, and loops, grows large, it may be possible to reduce the component count by using the same SPDs, JJs, and loops for multiple operations.

The circuits for various forms of synaptic plasticity presented in these Appendices are motivated qualitatively, and superior designs are undoubtedly possible. The circuit implementations for STDP, short-term plasticity, homeostatic plasticity, and metaplasticity are intended to convey the potential for diverse synaptic functionality achievable with superconducting optoelectronic circuits in the context of loop neurons. Because synaptic operations use few photons and fluxons, they are energy efficient. Scaling in complexity will likely be limited by fabrication challenges and device real estate.
	
\section{\label{apx:learningRate}Learning rate and memory retention}
To discuss synaptic update, it is helpful to define several parameters. We follow the conventions of Refs. \cite{fudr2005} and \cite{fuab2007}. We refer to the normalized synaptic weight as $w$, where $w=0$ corresponds to the minimum synaptic weight (in general not corresponding to a synaptic efficacy of zero), and $w=1$ corresponds to the maximum synaptic weight. The spacing between synaptic levels is denoted by $\alpha$. The total number of stable synaptic states between $w=0$ and $w=1$ is $1/\alpha$. Reference \onlinecite{fudr2005} defines a candidate plasticity event as ``the occurrence of a pattern of activity that could potentially lead to synaptic modification.'' These event occur at a rate $r$. The probability that one of these events is a candidate for strengthening (Hebbian) is $f_+$, and the probability that it is a candidate for weakening (anti-Hebbian) is $f_-$. The symbol $q$ denotes the ``size of the potentiation and depression modifications'' when synaptic update occurs \cite{fuab2007}. Synaptic strengthening occurs at a rate $qf_+r$, and weakening occurs at a rate $qf_-r$. The synaptic efficacy update rates as a fraction of the full synaptic efficacy range are given by $\alpha qf_+r$ and $\alpha qf_-r$. The initial signal\textendash to\textendash noise ratio of a memory upon storage is denoted by $S_0/N_0$, and is proportional to the number of synapses which have been modified by the memory. 

In the context of the circuits described here, $w$ is related to the synaptic bias current, $I_{\mathrm{sy}}$, which determines the synaptic efficacy. The synaptic efficacy is manifest physically as the current added to the NI loop during a synaptic firing event \cite{sh2018b}. In this work we have been treating $I_{\mathrm{sy}} = 1~\mu$A as the $w=0$ state of the synapse, and $I_{\mathrm{sy}} = 3~\mu$A as the $w=1$ state of the synapse. For the binary synapse of Figs. \ref{fig:synapticPlasticity_binaryCircuit} and \ref{fig:synapticPlasticity_binary}, $1/\alpha = 1$. For the multi\textendash stable synapse of Figs. \ref{fig:synapticPlasticity_supervisedCircuit} and \ref{fig:synapticPlasticity_supervised}, $1/\alpha$ was shown to be nearly 1000. $\alpha$ is determined by the synaptic storage loop inductance, and can take a wide range of values. The rate $r$ at which candidate Hebbian and anti-Hebbian events occur depends on the firing rates of the network, and this parameter is normalized out of analyses of memory retention times. The probabilities $f_+$ and $f_-$ also depend on network activity and in general cannot be relied upon to be precisely balanced \cite{fuab2007}. In the circuits described here, $f_+$ and $f_-$ can be engineered by changing the number of photons that are directed to the STDP receiver SPDs during each neuronal firing event. This number of photons can be much less than one so that a single photon is rarely directed for plasticity and synaptic update is infrequent. For example, we may operate in a mode wherein a pre\textendash synaptic neuron sends one photon to each downstream synaptic firing port, one photon to each downstream synaptic update strengthening port, and one photon to each downstream synaptic update weakening port during each neuronal firing event. The rate at which the post\textendash synaptic neuron sends photons to its own synaptic update ports then controls $f_+$ and $f_-$. This hardware-defined means of setting $f_+$ and $f_-$ can vary across a synaptic population. This approach to slow stochastic learning has the benefit of requiring few photons per neuronal firing event. A neuronal firing event would need to produce $3 k_{\mathrm{out}}$ photons, where $k_{\mathrm{out}}$ is the number of synaptic connections directed away from the firing neuron.

The size of synaptic modifications, $q$, is determined in the STDP circuit by the values of $I_{\mathrm{spd1}}$, $I_{\mathrm{spd2}}$, $I_{\mathrm{su}}^+$, and $I_{\mathrm{su}}^-$. As shown in Fig.\,\ref{fig:synapticPlasticity_Hebbian_1}, changing $I_{\mathrm{su}}$ changes the amount of current added to the SS loop during a synaptic update event, and therefore changes the current bias, $I_{\mathrm{sy}}$, which sets the synaptic weight during a synaptic firing event. 

Investigation of the limits of memory retention in the presence of ongoing plasticity \cite{fuab2007} reveals that memory lifetimes can be improved linearly with the number of stable synaptic states, $1/\alpha$. The expense is a decreased signal\textendash to\textendash noise ratio of stored memories, $S_0/N_0$. Reference \onlinecite{fuab2007} further discovered that synapses in which $q$ is a function of $w$ (``soft bounds'') performed well for extending memory storage times while maintaining high $S_0/N_0$. The circuits discussed in the present work can implement such soft bounds by inductively coupling $I_{\mathrm{su}}$ to $I_{\mathrm{ss}}$ so that $I_{\mathrm{su}}$ approaches some minimum value as the synaptic storage loop approaches saturation.

In addition to plasticity and multi\textendash stable synapses, power law memory retention is likely to make use of internal synaptic states \cite{fudr2005} that do not alter the efficacy of the synapse, but do affect the probability that a future Hebbian event will update the synaptic weight as well as affect the magnitude of that update, should it occur. This corresponds to states of the synapse with different values of $q$, but the same value of $I_{\mathrm{ss}}$. Circuit modifications that adapt learning rate in response to internal and external activity are referred to as metaplastic \cite{ab2008}. Using synapses with complex internal states allows for rapid incorporation of new information while maintaining stable, long\textendash term memories \cite{amfu1994,fudr2005,fuab2007,ab2008,khso2017}. A route to achieve metaplasticity in the circuits presented here is to vary $I_{\mathrm{su}}$, the current that determines the amount of synaptic shift during an update event. Thus, in the circuits presented here, we have $q(I_{\mathrm{su}})$. By changing $I_{\mathrm{su}}$ in time, plasticity can be present in an ensemble during a certain period of training, and then subsequently turned off, allowing those memories not to be corrupted by subsequent activity. $I_{\mathrm{su}}$ can by dynamically varied externally to implement supervised metaplasticity, or $I_{\mathrm{su}}$ can be modified by activity within the network using similar receiver circuits to those presented for STDP (see Appendix \ref{apx:metaplasticity}). Activity dependent modification not only of $I_{\mathrm{sy}}$, but also $I_{\mathrm{su1}}^+$ and $I_{\mathrm{su}}^-$ (see Fig.\,\ref{fig:synapticPlasticity_stdp}) is likely to provide mechanisms to adjust synaptic update rates to ensure the dynamic range of the synapses is matched to cortical activity \cite{bico1982,cobe2012} (see Appendix \ref{apx:homeostatic}).

Reference \onlinecite{fudr2005} elucidates that a network of heterogeneous synapses with a varying number of stable states does not outperform a network of binary synapses with multiple internal states, but a network of heterogeneous synapses with different numbers of stable states as well as multiple $q$ states was not investigated. This combination is likely to achieve the best of both worlds. The optoelectronic synapses of the present work have the opportunity to achieve spike\textendash timing\textendash dependent plasticity with a large number of stable levels as well as a large number of $q$ states affecting adaptation rate. If the goal is to achieve a learning system that can rapidly incorporate new information while retaining memories for a long time, neural systems must incorporate synapses that vary by different amounts and over different time scales. On a given neuron, or across an ensemble of neurons, a set of synapses may be heterogeneous in multiple capacities. The synapses may have a distribution in terms of number of stable states ($1/\alpha$), and they may have a distribution in learning rate, manifest in $q(I_{\mathrm{su}})$. Synapses that are updated infrequently and in small increments will store old wisdom. Binary synapses that switch readily bring fresh eyes. Neurons comprising primarily fresh eyes synapses bring fresh eyes to a network, and neurons comprising primarily synapses that were trained long ago and rarely change bring old wisdom. The ability to integrate information from a network of synapses with different learning rates trained at different times is advantageous for networks with optimal, power law forgetting rates. An ensemble of synapses is also likely to benefit from a diversity of short-term and homeostatic plasticity mechanisms, as discussed in Appendices \ref{apx:shortTerm} and \ref{apx:homeostatic}.

\bibliography{bibliography_modelingSOENs}

\end{document}